# Future Token Prediction - Causal Language Modelling with Per-Token Semantic State Vector for Multi-Token Prediction

Nicholas Walker

Iprova SA, Building 1, EPFL Innovation Park, 1015 Lausanne, Switzerland
`nwalker@iprova.com`

**Abstract.** Causal decoder-only transformer models used for generative language modelling, such as Generative Pre-trained Transformers (GPT), are trained to predict the next token in a sequence based only on its previous tokens. Despite this simple training objective, they have proved to be powerful AI tools. However, only predicting the next token results in top layer embedding vectors that are highly token-focused. There may be benefits in generating embedding vectors at each token position that better capture the overall meaning of longer sequences of future text.

Recent studies matching brain scans with deep language models suggest that humans also predict upcoming words when listening or reading but consider multiple future tokens rather than just one.

This research investigates a new pretraining method called Future Token Prediction (FTP). In FTP, a large transformer encoder generates top layer embedding vectors for each token position, which, instead of being passed to a language head, are linearly and expansively projected to a 'pseudo-sequence', which is cross attended to by a small transformer decoder to predict the next N tokens forward from that position in the sequence.

The top layer embedding vectors from FTP models exhibit distinct properties compared to those from standard GPT-like models, varying smoothly along a text sequence as measured by cosine similarity between adjacent tokens. Text generated by FTP models show improved topic coherence compared to standard GPT-like models trained with the same prediction perplexity for the next single token. The vectors are shown to better represent the topic of text based on the results of text classification examples. On a toy (but complex) coding problem, FTP networks produce significantly better results than GPT networks.

## 1   Introduction

Causal Autoregressive Large Language Models (LLMs) [1], the basis of developments in Generative AI, are transformer models [2] trained on the simple unsupervised task of predicting the next token from previous tokens in a sequence. Tokens are discrete elements of a sequential input, such as words or word parts, regions of an image or short sections of a signal. These models have proved to be very successful and do not appear to just act as plausible token generators but show behaviors beyond the statistical regurgitation of similar token sequences as occur in their training sets.

Such language models generate a top layer embedding vector at each token position, which is projected using a linear language head layer and a SoftMax to form a set of probabilities for each token in the defined model vocabulary. Text generation is performed by sampling tokens using these probabilities.

However, a single next token is not always representative of the semantics of the text needed to be generated, and as a result these models are prone to topic drift, particularly over longer sequences of token generation. Specific techniques [e.g. 3] can be used to address this drift but such approaches do not address the underlying defects of a learning task based only on one token ahead prediction [4].

In [5], the authors found linear mappings from the activations of transformer language models to human functional magnetic resonance imaging brain signals for words occurring in the future, but the prediction score was maximized with a prediction of up to 8 to 10 future words.

Pal et al. [6] find that the different language model transformer layer embeddings are able, to some extent, to predict future tokens beyond one ahead. Therefore, a suitable research objective would be to determine if there is a reasonable way to set up modified language models where the prediction extends to the next 8 to 10 tokens, whilst maintaining low perplexity for next token prediction and therefore keeping the ability to perform simple (one token ahead) autoregressive token generation, which is the basis of Generative AI.

## 2 Previous Work

Researchers have explored several approaches which appear to offer the capability of predicting multiple tokens ahead.

### 2.1 Multi-token prediction 'in one go'

A simple method for predicting multiple future tokens is to either use multiple linear language heads (one per future token) to project the entire top layer embedding vector to more than one array of future token probabilities or to split the embedding vector into segments each of which projects through the same linear language head to form the multiple future token probability arrays. For example, Lando22[1] has experimented with a GPT-like model that predicts the next 3 tokens from a split top level embedding but reports only poor results. A more successful work is that of Gloeckle et al. [7] who find that projecting the entire top layer embedding with 4 different linear projection language heads to the next 4 tokens improves the performance of LLMs for coding.

### 2.2 Seq2Seq models

Another obvious approach to multi-token prediction is to use a Seq2Seq transformer model, with an encoder and a decoder, with the decoder predicting the next N tokens. In Seq2Seq models, no specific per token position vector is passed to the decoder - the decoder has access, via cross attention, to all the top layer embeddings of the encoder sequence. Such approaches have been examined in XLNet [8] and T5 [9]. In this approach the decoder's previous token context – all the top layer token embeddings from the encoder – is the same as the encoder context so this approach does not 'distill' the context for the decoder into a single embedding vector.

### 2.3 Masked Language Models

Masked language models also have an ability to fill in specific masked tokens (e.g. Bert [10]) or spans of tokens (SpanBert [11]) and there is no reason why more than one masked token cannot be passed to the models at the end of the prompt for the model to infer multiple future tokens. A range of approaches have examined masked language models for multi-token prediction [12, 13, 14]. However, all these approaches have not proved to be very successful at left to right text generation. Again, the previous context of a token is not distilled into a single embedding vector – the model attends to all previous tokens to determine the future tokens.

### 2.4 Modified Transformer Decoder

ProphetNet [15] is a Seq2Seq pretraining model which introduces a novel self-supervised objective named future n-gram prediction and an n-stream self-attention mechanism to do this. Instead of the optimization of one-step ahead prediction in traditional sequence-to-sequence model, ProphetNet is optimized by n-step ahead prediction which predicts the next n tokens simultaneously based on previous context tokens at each time step. The future n-gram prediction explicitly encourages the model to plan for multiple future tokens and prevent overfitting on strong local correlations. The model architecture is based on the original Transformer but replaces the "standard" self-attention mechanism in the decoder by a main self-attention mechanism and a self and n-stream (predict) self-attention mechanism for predicting multiple future tokens.

### 2.5 Recurrent Neural Networks for Multi-token decoders from embedding vectors

Somewhat similarly to the work of this paper, in [12] the authors also propose taking the top layer token embedding vector of a masked transformer encoder and passing this to a Recurrent Neural Network (GRU [17]) to reconstruct a short multi-token sequence. However, the use in their work is not generic for future text tokens but only applied to a subset of tokens representing masked multi-word expressions.

---

[1] https://github.com/lando22/GPT-3T

# 3   Overview of Approach

Unlike Recurrent Neural Networks, in a transformer architecture there is no single hidden vector which captures the full past token context to be used in subsequent text generation. Instead, this achieved by interrogating the past context via the attention mechanism operating on the embeddings of all previous tokens. Our goal instead is to pass the top layer token embedding from the encoder as the sole past context for a decoder, operating at each token position, to predict the next N tokens ahead from that position. In the description here, the normal GPT *decoder* is termed an *encoder* to differentiate it from the multi-token generation decoder in the FTP model.

We implement an *encoder* which attends to its previous tokens, and outputs at its top layer an embedding vector for each input token. A standard GPT would apply the language head to this embedding vector. Instead, we supply this output embedding (for each token in the sequence), used in a somewhat similar manner to the hidden vector in a recurrent neural network language model [16, 17], to a transformer *decoder*, one decoder instance for each token position. The decoder cross attends to this single embedding vector (only) and self-attends to its own previous tokens (but no previous tokens in the initial sequence) to generate the next N tokens. Some of the capability for future token generation is then carried by the decoder acting as a short (N-token) causal sequence language model, but most of this predictive capability is supplied by the embedding vector. We choose to keep the number of decoder layers small (in this work 3) as compared to the encoder (here, 12) so the embedding vector must be preferentially used rather than just using the decoder as a language model. At training, we use teacher-forcing in the decoder, supplying as input the N tokens starting with the last token in the encoder input (for that top layer embedding) and as target the next N tokens one position ahead of the decoder inputs. At autoregressive inference we can seed the decoder with the current token (and not any previous tokens) and then sample and use the generated tokens.

However, this leaves an issue for a transformer decoder in that we have only a single token embedding (of the encoder embedding dimension size), i.e. a sequence length of one, rather than an actual sequence of such embeddings (such as might be generated by a transformer encoder in a Seq2Seq model). This means the cross-attention mechanism in the decoder, whose purpose is to variably attend to multiple tokens, makes little sense.

To address this, a linear layer is used to expansively project the top layer token embedding vector, of dimension 1 by *Dim* (where *Dim* is the decoder embedding dimension, generally the same as that of the encoder) into a 'pseudo-sequence' whose dimensions are *Seq* by *Dim*. *Seq* is a short, fixed chosen sequence length (e.g. 12). In this way, the MLP can place information from the embedding vector along a learned sequence which can be variably attended to by the decoder cross attention mechanism.

| Model parameters | Value |
|---|---|
| Encoder layers | 12 |
| Encoder hidden dimension | 768 |
| Encoder Attention heads | 12 (16 for code) |
| Encoder MLP dimension | 2304 |
| Encoder token length | 1024 |
| Total Encoder parameter count (excluding Embedding/LM head) | 92.03M |
| Decoder projection *Seq* value | 12 |
| Decoder linear projection parameter count | 7.08M |
| Decoder layers | 3 |
| Decoder dimension | 768 |
| Decoder Attention heads | 12 (16 for code) |
| Decoder MLP dimension | 2304 |
| Decoder token length | 8 |
| Total Decoder parameter count (excluding Embedding/LM head) | 30.09M |
| Total GPT parameters (excluding Embedding/LM head) | 92.03M |
| Total FTP parameters (excluding Embedding/LM head) | 129.21M |

**Table 1.** Parameters for GPT (encoder only) and FTP (encoder, projection layer and decoder) used in this work.

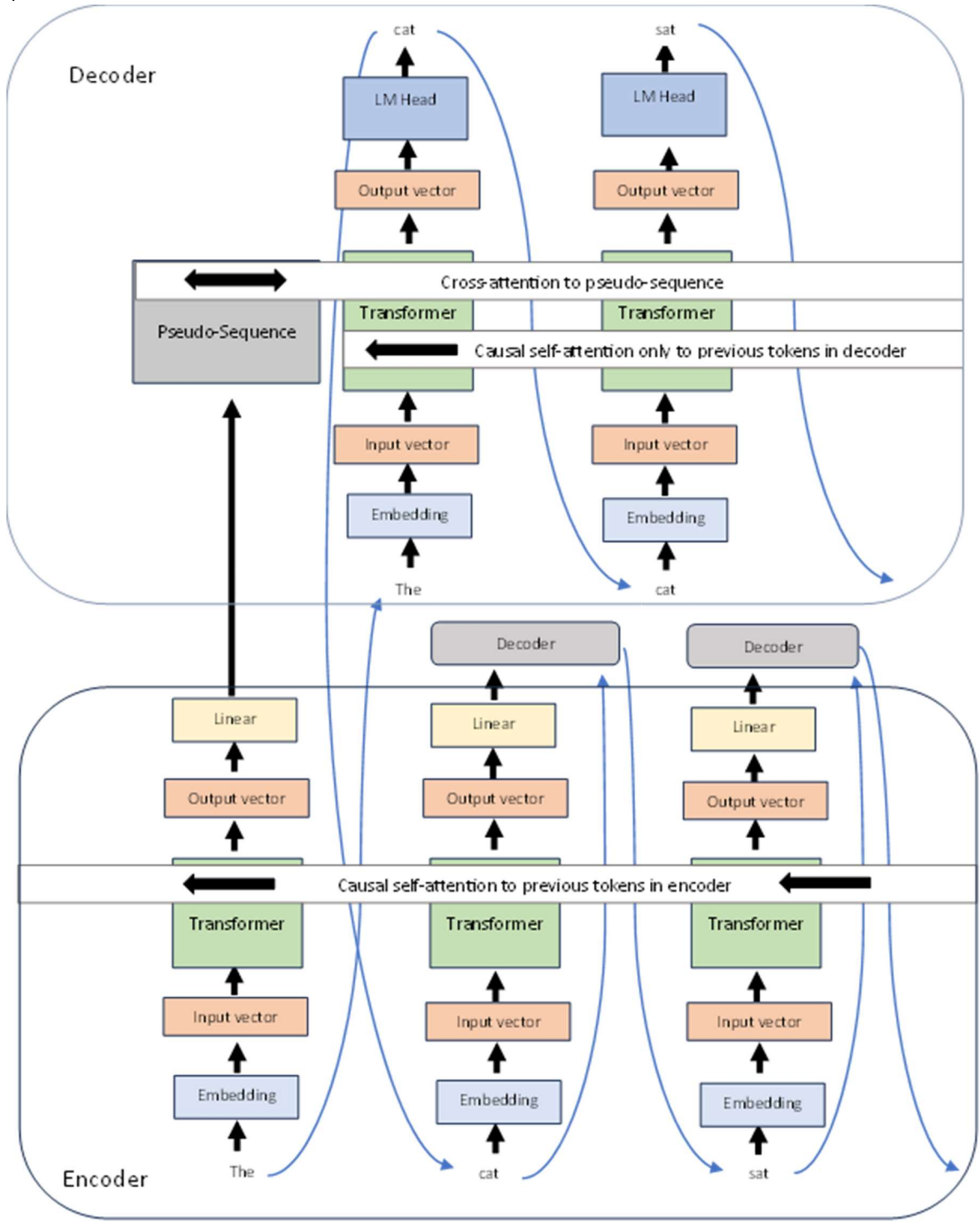

**Fig. 3.** FTP model: A transformer encoder produces a top layer token embedding vector at each token position which is processed by an MLP, here simply a linear layer, to produce a *Seq* by *Dim* 'pseudo-sequence' array, where *Seq* is a chosen sequence length (12 in this work) and *Dim* is the decoder embedding dimension (generally the same as the encoder embedding dimension). A transformer decoder is given the same first input token as at that token position in the encoder (plus during training, the following N-1 tokens), performs causal self-attention on previous token positions in the decoder and cross attention on the entire pseudo-sequence. During GPT-like inference only the first token from the decoder is output as the predicted next token. All embedding and LM head weights are shared. Transformers consist of multiple layers. Cross attention in all decoder layers is to the same pseudo-sequence.

During training we supply the encoder with the tokenized sequence. At each top layer encoder position associated with this input sequence, we generate the 'pseudo-sequence' for the decoder to cross attend to, plus as input to the decoder, the following N tokens starting at the token associated with that position, and a training target, which is the next N tokens starting from the following token (that is one ahead of the decoder input tokens).

To focus the learning process on the earlier tokens, token losses at the decoder are exponentially down weighted by their distance ahead, with a chosen gamma (here 0.8). The loss on the first token ahead is scaled by 1.0, the second by gamma, the third by gamma squared, the fourth by gamma cubed and so on.

Such Future Token Prediction (FTP) models can generate text autoregressively, one next token at a time, in the same manner as GPT models, without autoregressive generation in the decoder. Alternatively, more complex generation strategies can be used, where token probabilities from the decoder further ahead are also considered.

## 4 Method

### 4.1 Models

A pre-layer norm GPT2-like model was implemented using as a basis the Karpathy NanoGPT code[2] but with the learned positional embeddings replaced by XPOS [18] rotational position embeddings[3] added at each layer. Bias was set to False. The transformer MLP used is the SwiGLU MLP [19], with an expansion factor of 3.

A medium scale GPT2 model (12 layers, 12 heads, 768 embedding dimensions) was used for the encoder and a small GPT2 model with additional cross attention (3 layers, 12 heads, 768 embedding dimension) was used for the decoder, again with XPOS rotary embeddings used at each layer. The decoder also has learned absolute position embeddings added to the token embeddings prior to the first layer. The encoder token embedding weights, decoder token embedding weights and decoder language model head weights were shared.

The standard GPT2-like model used had the same parameters and structure as the FTP encoder but with a language model head (shared with the token embeddings) applied to the top layer embedding vector and no decoder.

The number of predicted future tokens, N, was chosen as 8 and the *Seq* parameters (for the 'pseudo-sequence' projection) as 12.

### 4.2 Training

The OpenWebText[4] text collection was used for training. This was tokenized using the GPT2 tokenizer and randomly sampled from during training, as per the Karpathy code. A data retrieval function was implemented to return the 1024 length token sequence for the encoder along with the aligned 1024 * 8 decoder input sequences and 1024 * 8 decoder target sequences (which are one token ahead of the decoder input sequences).

AdamW was used for optimization with an accumulated batch size of 500, encoder context length of 1024, decoder context length of 8. Approximately 12 million encoder sequences (each of 1024 tokens) were used in training, with a linear warmup of 500,000 sequence presentations and cosine learning rate reduction starting at learning rate of 4e-4. Weight decay was 0.1. Both models were trained to a final validation loss of 3.0. Neither model had fully converged at this point.

Losses on the targets were generated using cross-entropy with one-token ahead for the GPT-like model and with the decoder target tokens for FTP model (1-8 tokens ahead), As stated, in FTP, future token losses for each future token from the decoder were discounted using an exponentially decreasing weighting factor with a defined gamma according to the future distance. A gamma value of 0.8 was selected.

### 4.3 Inference

At inference time, instead of the linear language model head of the standard GPT2 model, we pass a single token input to the decoder (which is the last input token in the encoder sequence). This will generate a single top layer decoder embedding, which is projected to the logits for the vocabulary by the decoder language model head to allow sampling the next token. Therefore, for equivalent behavior to a standard GPT-like model, we do not need to run the decoder autoregressively. However, by training the encoder in this way, we hope to have biased the embedding vector to try to predict (with progressively less confidence as we go further into the 'future') the possible future N token distribution.

---

[2] https://github.com/karpathy/nanoGPT
[3] https://github.com/lucidrains/rotary-embedding-torch
[4] https://skylion007.github.io/OpenWebTextCorpus/

By ensuring the decoder is significantly smaller than the encoder, the computation and memory overhead is relatively small particularly as we increase the encoder size.

### 4.4 Lookahead inference

The decoder constitutes a light weight means of performing autoregressive lookahead. Different methods might be considered for how this lookahead can be incorporated for next token sampling, but a simple one which preserves the ability to probabilistically sample next tokens for autoregressive generation is the following:

1. Non-autoregressively generate the predictions for the next token using the decoder with the single last sequence token as input as in section 4.3
2. Select a number K of 'top' next tokens, using topK sampling from the next token logits, converting their logits into probabilities (limited to those top tokens). Each topK token is then associated with an initial log probability in a K-length array (these would be the probabilities we sample from in the case of simple one-token ahead prediction in the FTP model).
3. For each of these tokens added to make a length 2 prompt (following the last sequence token), perform a further greedy autoregressive generation using the decoder up to a defined look ahead distance L (e.g. performed in parallel in a batch of size K). At each next token production stage convert the logits into a log probability for the single next (highest probability) token and add this into the running cumulative probability value for each of the original selected tokens, correctly down weighted by the gamma factor for that distance ahead.
4. At the end normalize the cumulative probability array over the K tokens to a sum of 1.
5. Finally sample the initial selected K tokens but using the calculated combined greedily generated sequence probabilities rather than the initial (one token ahead) probabilities.

## 5 Results

### 5.1 Similarity of embedding vectors over a sequence

In a standard next token prediction transformer, the embedding represents the token to be generated. Therefore, there should not be any specific relationship between nearby embeddings. In FTP, we expect only a small variation in value since the embedding represents a 'future prediction' which should smoothly change.

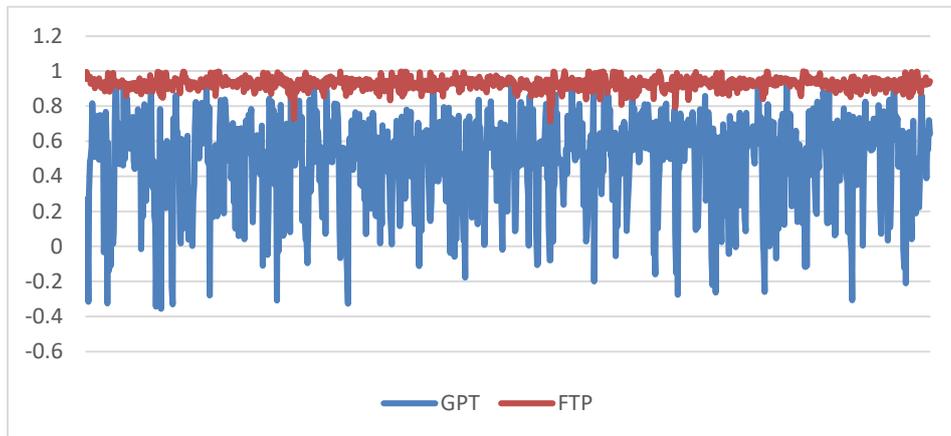

**Fig. 3.** Example adjacent token top layer embedding vector cosine similarity scores over a 1024 token sequence randomly selected from the validation set for GPT-like Language Model (GPT) and Future Token Prediction (FTP) Language Model. The FTP top layer embeddings have a significantly higher similarity and lower variation.

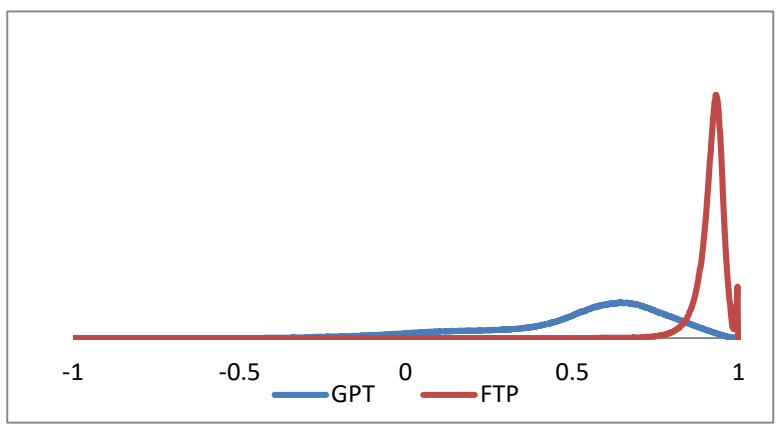

**Fig. 4.** A comparison between the 1-token separation cosine similarity distributions for GPT-like and FTP models.

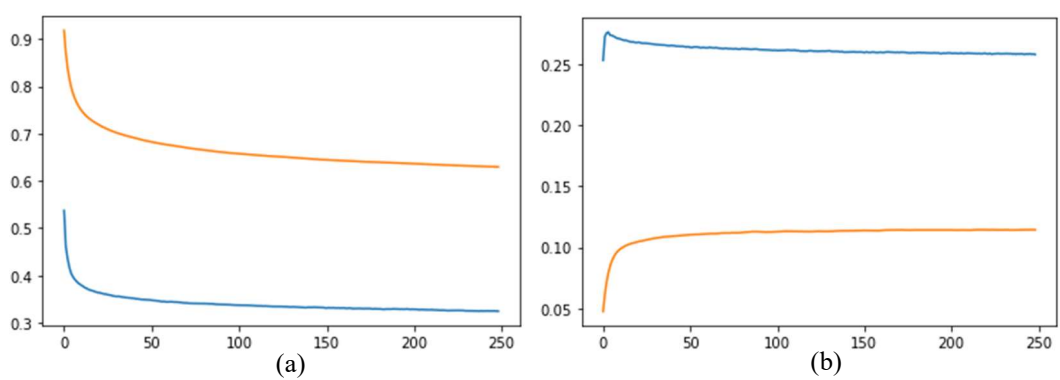

**Fig. 5.** shows (a) mean and (b) standard deviation of cosine similarities between token embeddings for different token separations (up to 250) for GPT (blue) and FTP (orange). GPT 'infinite' separation mean (std) was 0.153 (0.23) and FPT 0.402 (0.118).

### 5.2 Determining the predictive ability of top layer embeddings for future token prediction.

The trained FTP model was compared to the GPT model using probes to determine the accuracy of predicting future tokens, with perplexity numbers noted from a validation set.

**For the GPT2-like model:**
   a. The native 1 token ahead perplexity of the model using its original Language Model head was noted.
   b. 2-layer MLPs (different for each future token) with expansion of 4 and Gelu non-linear function followed by a linear Language Model head were trained using the training set to predict future token perplexity for look ahead of 2 to 5 tokens using the top layer token embeddings with the GPT model frozen.
   c. A transformer decoder with the same parameters as the decoder used for the FTP model was trained using the training set to autoregressively predict future tokens for look ahead of 10 tokens using top layer embeddings in a similar manner to the future token prediction model decoder, and the perplexity for tokens 1-5 noted. Weightings for tokens ahead were set to be the same exponentially decreasing function as in the FTP model training. The weights of the GPT2-like model were frozen.

**For the FTP model:**
   a. 2-layer MLPs (different for each future token) with expansion of 4 and Gelu activation function followed by a linear Language Model head were trained to predict future token perplexity for look ahead of 1 to 5 tokens using top layer embeddings with the FTP model frozen.
   b. The original model decoder future token perplexity for tokens ahead by 1 to 5 was noted.

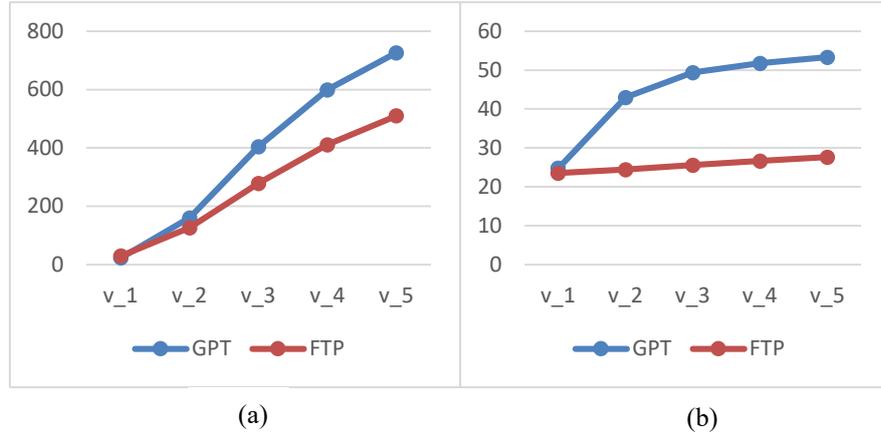

**Fig. 6.** Perplexity for future token prediction. (a) trained MLP probe (different for each future token) (b) teacher forced trained transformer decoder (as in FTP model). v_1 is one token ahead, v_2 is 2 tokens ahead and so on.

The FTP model performs considerably better at predicting future tokens than the GPT-like model, as expected.

### 5.3   Text generation quality and topic adherence

Various automated metrics are available for the measurement of text quality and diversity. We most wish to evaluate text semantic match to the actual following text section.

1,000 random 200 token length selections are made from the validation file filtered to remove sequences containing an <endoftext> token. The first 100 tokens are used as the 'prompt', and the subsequent 100 tokens taken as the 'actual continuation'. Using the prompt, the LMs are used to autoregressively generate 100 additional tokens as 'generated continuations' (with selection of the <endoftext> token suppressed).

For each prompt, 10 generated continuations were made, with sampling based on topK of 100 and a temperature of 1.0. BERT scores [20] were used to evaluate the semantics of the text match.

The metric is calculated between:
1. actual prompt and a) actual continuation and b) generated continuations
2. actual continuation and generated continuations

The mean metric scores for the 10 equivalent continuations over the 1,000 examples are determined for the two LM types. The Hugging face 'evaluate' module[5] was used to perform the analysis.

| **Continuation compared to actual prompt** | Actual | GPT | FTP |
|---|---|---|---|
| BERT Score precision | 0.76459 | **0.76090** | 0.76062 |
| BERT Score recall | 0.76433 | 0.75688 | **0.75691** |
| BERT Score f1 | 0.76435 | **0.75874** | 0.75865 |
| **Continuation compared to actual continuation** | | GPT | FTP |
| BERT Score precision | | 0.74558 | **0.74833** |
| BERT Score recall | | 0.74195 | **0.74474** |
| BERT Score f1 | | 0.74363 | **0.74641** |

**Table 2.** Generative text matching quality metrics for GPT and FTP models.

The GPT and FTP cosine similarity scores for generated continuations compared to prompts are comparable and less than the scores for actual continuation compared to prompt. The FTP scores for cosine similarity of generated continuations compared to actual continuations are increased over the GPT scores.

---

[5] https://huggingface.co/spaces/evaluate-metric/bertscore

**5.4   Text Classification from frozen model embeddings**

Text classification was performed based on outputting a text sequence set of embeddings from frozen GPT and FTP models, forming a mean of the embeddings (except the first token) and using a 2-layer MLP with GELU non-linearity plus a final linear layer to learn to classify the text. The IMDB review sentiment dataset[6], Movie genre dataset[7] and Amazon review[8] sentiment datasets were used. A DistilBERT text classifier[9] was also fully fine-tuned to provide a 'best possible result' metric.

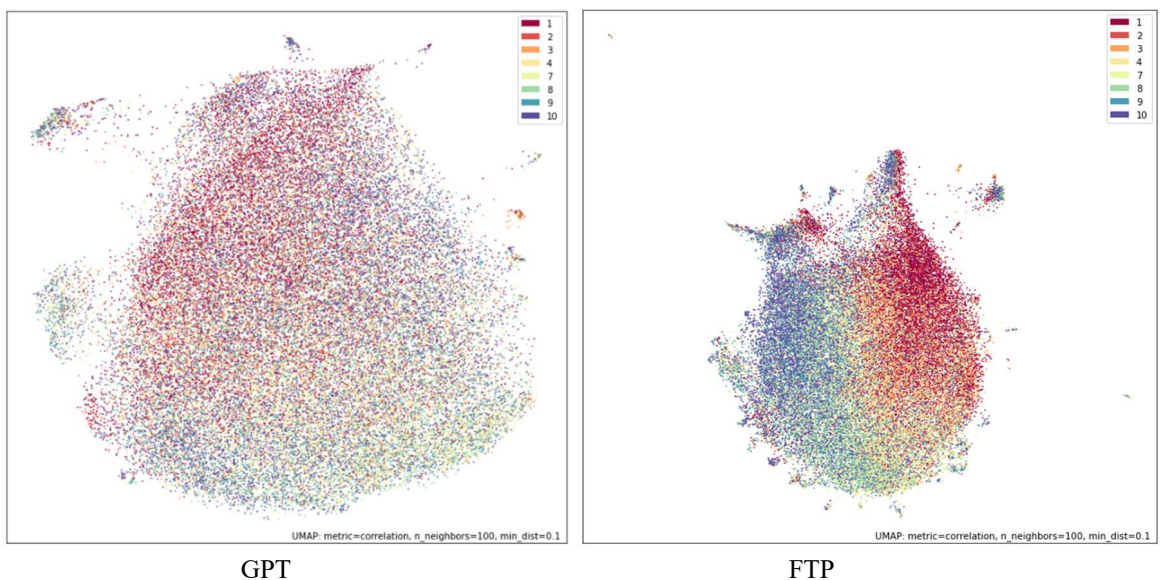

GPT                                                                 FTP

**Fig. 7.** UMAP plot of mean vectors for the IMDB movie sentiment analysis dataset – the FTP plot shows a cleaner demarcation of vector positions.

| Dataset | Network | Validation Loss | Validation Accuracy |
| --- | --- | --- | --- |
| IMDB – all categories | GPT | 1.4278 | 0.43776 |
| IMDB – all categories | FTP | **1.3827** | **0.45264** |
| IMDB – all categories | DistilBERT | 1.329033 | 0.497040 |
| IMDB – binary | GPT | 0.25865 | 0.89396 |
| IMDB – binary | FTP | **0.22054** | **0.91180** |
| IMDB – binary | DistilBERT | 0.210118 | 0.928840 |
| Movie genre | GPT | 1.235614 | 0.616771 |
| Movie genre | FTP | **1.201091** | **0.628875** |
| Movie genre | DistilBERT | 1.133098 | 0.660498 |
| Amazon reviews | GPT | 0.181023 | 0.930243 |
| Amazon reviews | FTP | **0.15364** | **0.943135** |
| Amazon reviews | DistilBERT | 0.097692 | 0.970615 |

**Table 3.** Validation loss and accuracy for classification datasets. For GPT and FTP only a final MLP is trained using mean top layer embedding vectors for a top layer embedding sequence from frozen models. The DistilBERT text classification models are fully fine-tuned.

---

[6] https://www.kaggle.com/datasets/lakshmi25npathi/imdb-dataset-of-50k-movie-reviews
[7] https://www.kaggle.com/datasets/khushipitroda/movie-genre-detection
[8] https://www.kaggle.com/datasets/bittlingmayer/amazonreviews/data
[9] https://huggingface.co/docs/transformers/en/tasks/sequence_classification

A simple polling (mean) on the top-level token embeddings results in a representation vector delivering consistently better accuracy and lower losses for FTP compared to GPT models. The results are worse than SOTA as expected, but the models themselves were not fine-tuned (only the MLP projection layers).

### 5.5 Topic adherence in long generated sentences using sentence embedding vector matching

A set of queries topics are converted into prompts and 800 tokens of generated text are generated, for GPT and for FTP with up to 8 tokens of lookahead. 10 queries were written and 10 examples of generation for each query used.

The generated text is tokenized into sentences using the NLTK sentence tokenizer and the sentence embedding vectors from a SentenceTranformer[10] (all-MiniLM-L12-v2) are compared to the embedding vector of the query using cosine similarity. A mean is taken over all the sentence cosine similarities (excluding the first which includes the query and the last, which may only be partially formed).

The average cosine similarity of generated sentences to prompt was 0.2226 for GPT and 0.2502 for FTP.

The FTP model is consistently better at achieving a level of topic adherence over long generated text compared to GPT. Increasing the lookahead using the approach described in section 4.4 further improves the results, with good values obtained at around 5 look-ahead tokens.

Examples are given in the appendix of a non-cherry-picked set of 800 token generations from GPT, FTP with next token prediction and FTP with a 5 token probability look ahead.

### 5.6 Coding Language Model

Based on the work of Jin and Renard [21], transformer models for inferring a program for progressing start grid worlds into stop grid worlds were trained. Five 8 by 8 start and stop grids are first presented as tokens in the text string, followed by the 6-10 length program for transforming the start into the stop grids. The grids consist of obstructions (including all the outer edge cells) plus cells which are marked with a score between 0 and 10. A 'turtle-like' entity is placed on a non-obstructed cell, pointing in one of the 4 cardinal directions. The start grids and programs are randomly generated with the stop grid generated from running the program on the start grid.

A 6-10 (training) or 1-10 (testing) step length program is appended to the grid tokens, consisting of the possible values: move (in current entity direction, no move if cell to move to is an obstruction), rotate left, rotate right, mark (increase value of current position, maximum 10) and unmark (decrease value of current position, minimum 0). Finally, an <EOS> token is appended to be followed by pad tokens. The sequences then consist of: the grids forming 640 tokens, then the program additional tokens (maximum 10 tokens), the <EOS> token and all the strings are then padded with a <pad> token up to a length of 662 tokens.

One million programs are generated for training and 10,000 for testing, all programs for both training and testing being unique.

During evaluation, following supplying the 640 token 'prompt' of the five start and stop grids, the greedily generated programs are checked for the number of unique programs generated and their correctness (as a proportion of the total).

Small GPT and FPT models, identical to the language models presented above (12-layer encoder, 3-layer decoder for FTP) except for increasing the attention heads in encoder and decoder to 16, were trained for 12 epochs (FTP model runs were truncated at 9 epochs as no further improvement was found).

---

[10] https://sbert.net/docs/sentence_transformer/pretrained_models.html

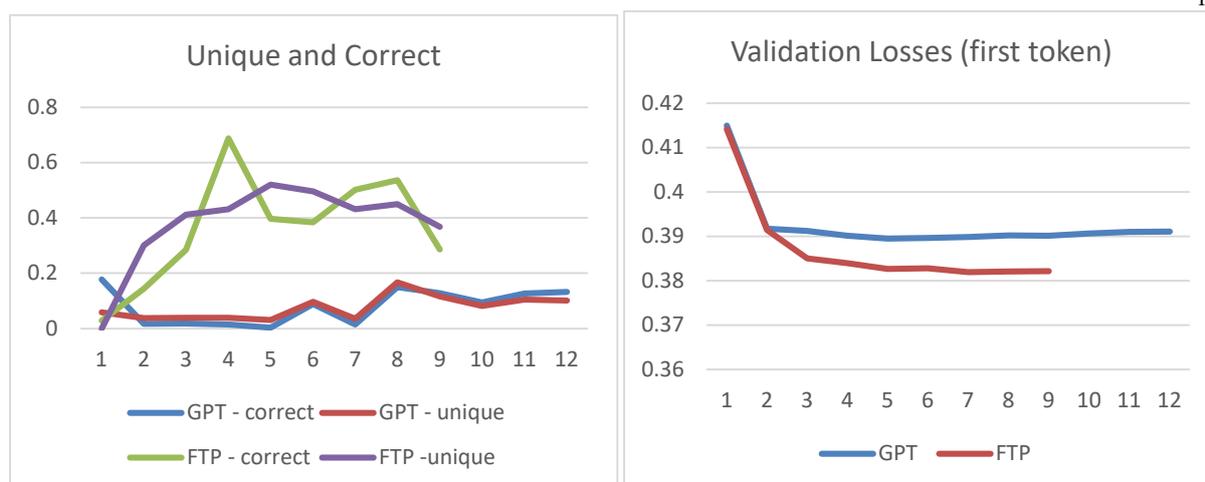

**Fig. 7.** Proportion of unique and correct programs generated by greedy decoding following 5 start and stop grids plus overall validation first token losses for GPT and FTP models.

The FTP models consistently achieved significantly better results than the GPT models.

## 6   Conclusions

A new approach to autoregressive language models and their training, termed a Future Token Prediction model or FTP, is proposed. In these models the initial causal language model encoder generates, at each token position, a vector embedding which, instead of being used with an LM head to generate the probabilities from which the next token is sampled, is passed to a small autoregressive LM decoder to output the next N tokens (in this work N was chosen as 8). The embedding vector from the encoder is first passed to a linear layer to expansively generate from it a learned 'pseudo-sequence' (here of length 12) of decoder embedding dimension, which is the 'sequence' cross-attended to by the decoder. The cross-entropy training losses on the future tokens are exponentially down weighted according to their future distance by a gamma factor.

Two similar LMs were trained based on a modified GPT2 model, one a standard LM and one the Future Token Prediction LM. The resulting embedding vectors from the future token prediction LM have very different properties to the embedding vectors from a standard LM, changing smoothly along a text sequence except where the sequence is hard to predict. The future token LM is better able to predict tokens beyond one ahead compared to the standard LM, and metrics of text quality show improvements, specifically of generated text semantics compared to actual text semantics, compared to a standard GPT LM with the same one token ahead prediction perplexity.

In a coding example, the FTP model provides significantly better results than the GPT model.

By forcing a LM to generate a more consistent and token-independent embedding of future text, it appears that the LM better learns internal models which represent a generally smoothly changing 'world state' from which the next token(s) is determined.

The number of parameters in small models, along with the training times, are certainly increased, but this relative difference should reduce as the model encoders increase in size.

Further research will extend this approach to significantly larger models trained on Iprova's creativity-related text datasets, which are currently being compiled.

## References

1. Radford A., Narasimhan K., Salimans T., Sutskever I.: Improving language understanding by generative pre-training. OpenAI Blog. (2018). https://cdn.openai.com/research-covers/language-unsupervised/language_understanding_paper.pdf

2. Vaswani A., Shazeer N., Parmar N., Uszkoreit J., Jones L., Gomez A. N., Kaise L., Polosukhin I.: Attention Is All You Need, 31st Conference on Neural Information Processing Systems (NIPS 2017), Long Beach, CA, USA. (2017). https://arxiv.org/abs/1706.03762
3. Xu N., Zhou C., Celikyilmaz A., Ma A.: Look-back Decoding for Open-Ended Text Generation. Proceedings of the 2023 Conference on Empirical Methods in Natural Language Processing, pages 1039–1050 (2023). https://arxiv.org/pdf/2305.13477
4. Bachmann G., Nagarajan V.: The Pitfalls of Next-Token Prediction, Proceedings of the 41st International Conference on Machine Learning, PMLR 235:2296-2318, (2024). arxiv.org/pdf/2403.06963.pdf
5. Caucheteux C., Gramfort A., King J-R.: Evidence of a predictive coding hierarchy in the human brain listening to speech, Nature Human Behaviour, volume 7, pages 430–441 (2023). https://www.ncbi.nlm.nih.gov/pmc/articles/PMC10038805/
6. Pal K., Sun J., Yuan A., Wallace B.C., Bau D.: Future Lens: Anticipating Subsequent Tokens from a Single Hidden State. Proceedings of the 27th Conference on Computational Natural Language Learning (CoNLL), pages 548–560, Singapore. Association for Computational Linguistics (2023). https://arxiv.org/pdf/2311.04897
7. Gloeckle F., Idrissi B. Y., Rozière B., Lopez-Paz D., Synnaeve G.: Better & Faster Large Language Models via Multi-token Prediction. (2024). https://arxiv.org/pdf/2404.19737
8. Yang Z., Dai Z., Yang Y., Carbonell J., Salakhutdinov R., Le Q.V.: XLNet: Generalized autoregressive pretraining for language understanding. In Advances in Neural Information Processing Systems (NeurIPS) (2019). https://arxiv.org/abs/1906.08237
9. Raffel C., Shazeer N., Roberts A., Lee K., Narang S., Matena M., Zhou Y., Li W., Liu P.J.: Exploring the limits of transfer learning with a unified text-to-text transformer. Journal of Machine Learning Research 21 (2020) 1-67 (2019). https://arxiv.org/pdf/1910.10683
10. Devlin J., Chang M-W., Lee K., Toutanova K. Bert: Pre-training of deep bidirectional transformers for language understanding. In Proceedings of NAACL-HLT (2019), pages 4171–4186. (2019). https://arxiv.org/pdf/1810.04805
11. Joshi M., Chen D., Liu Y., Weld D.S., Zettlemoyer L., Levy O.: Spanbert: Improving pre-training by representing and predicting spans. Transactions of the Association for Computational Linguistics, vol. 8, pp. 64–77 (2020). https://arxiv.org/pdf/1907.10529
12. Kalinsky O., Kushilevitz G., Libov A., Goldberg Y.: Simple and Effective Multi-Token Completion from Masked Language Models. In Findings of the Association for Computational Linguistics: EACL 2023 (2023). https://aclanthology.org/2023.findings-eacl.179.pdf
13. Du Z., Qian Y., Liu X., Ding M., Qiu J., Yang Z., Tang J.: GLM: General Language Model Pretraining with Autoregressive Blank Infilling. Proceedings of the 60th Annual Meeting of the Association for Computational Linguistics, Volume 1: Long Papers, pages 320 – 335 (2022). https://arxiv.org/pdf/2103.10360
14. Bao H., Dong L., Wei F., Wang W., Yang N., Liu X., Wang Y., Piao S., Jianfeng Gao J., Zhou M., Hon H-W.: UNILMv2: Pseudo-Masked Language Models for Unified Language Model Pre-Training. Proceedings of the 37th International Conference on Machine Learning, Online, PMLR 119, (2020). https://arxiv.org/abs/2002.12804
15. Qi W., Yan Y., Gong Y., Liu D., Duan N., Chen J., Zhang R., Zhou M.: ProphetNet: Predicting Future N-gram for Sequence-to-Sequence Pre-training. Findings of the Association for Computational Linguistics: EMNLP 2020, pages 2401–2410 (2020). https://arxiv.org/abs/2001.04063
16. Hochreiter S., Schmidhuber J.: Long Short-Term Memory. Neural Computation (1997) 9 (8): 1735–1780. (1997). https://deeplearning.cs.cmu.edu/S23/document/readings/LSTM.pdf
17. Chung J., Gulcehre C., Cho KH., Bengio Y.: Empirical Evaluation of Gated Recurrent Neural Networks on Sequence Modeling, NIPS 2014 Deep Learning and Representation Learning Workshop (2014). https://arxiv.org/abs/1412.3555
18. Sun Y., Dong L., Patra B., Ma S., Huang S., Benhaim A., Chaudhary Y., Song X., Wei F.: A Length-Extrapolatable Transformer. In Proceedings of the 61st Annual Meeting of the Association for Computational Linguistics (Volume 1: Long Papers), pages 14590–14604 (2023). https://arxiv.org/abs/2212.10554
19. Shazeer N.; GLU Variants Improve Transformer, https://arxiv.org/pdf/2002.05202
20. Zhang T., Kishore V., Wu F., Weinberger K.Q., Artzi Y.: BERTScore: Evaluating Text Generation with BERT, ICLR 2020 (2020). https://arxiv.org/abs/1904.09675
21. Jin C., Rinard M.: Emergent Representations of Program Semantics in Language Models Trained on Programs, Proceedings of the 41st International Conference on Machine Learning, Vienna, Austria. PMLR 235, (2024), https://arxiv.org/pdf/2305.11169v3

# Appendix: Examples of long text generation

**GPT (one token ahead)**

**How is it possible for a computer system to be truly creative? This requires that** a computer learn that it is not a human being.

That may not come easy when computers can't, by definition, work in a vacuum (even if they can). But not so with us. It has been nearly a decade since Microsoft began designing this software.

In its case, the real breakthrough in computing was the creation of the Kinect. Then, thanks in large part to Kinect, Xbox got some attention from researchers and companies for its real-time video-sharing capabilities. But at the time, Microsoft had its Kinect as a kind of backup computer. Now that it has become a part of every household, it's not a "mater" at all.

But now, thanks to Mr. Kinect, now we can all get together and share "a million instant videos" instead of just sitting down. He is excited about this, too. "At some point, it will probably become really important to connect the two projects," he says.

Then Microsoft came along.

It gave up the personal computing race. "In the three decades since (Microsoft's) last successful creation of software, we have been so focused on the personal computing market that we have lost the personal computing revolution," Ms. McIlwaine says.

Microsoft is hoping to have Kinect's first feature -- a webcam -- ready for mass production right away sometime this year if, say, it can prove to be a significant improvement over the Kinect.

You see, the Kinect is not quite as deep as Microsoft thought it was -- more like a mouse. (Perhaps even more importantly -- it doesn't have to do extra work.) How far are we talking here? Well, Microsoft has recently been considering whether to start working with large companies. And guess what? How many games are you looking at now? These days, there are about a dozen. Which is great. But the thing about the Kinect is amazing. At its closest to 0,000 feet, it has to pull in 3,000 feet. If the Kinect could ever be able to do it, we probably would not need that much.

Photo

There's no way that Kinect made its way across the United States in any direction from 1995. But that doesn't mean it isn't useful for the production of Windows. Where it probably doesn't make the hardware for the PC, Kinect has the potential to create much-needed hardware for homes and workplaces.

The "nautilus" technology that Microsoft was thinking about, the "smartphones" and "smart-home" (something "smart") to assist devices is something that has been around for a century.

The problem is that the Kinect is, perhaps purposefully, not as huge as it seems. It was built to make simple video-sensing tasks a lot easier than they need to be."

And Microsoft has no control over those tasks when they are made. "People use video games and say, 'Oh, that's my first job,' " Ms. McIlwaine says. "But they still need to know that those are real jobs."

So it's not easy to use Microsoft's Kinect, but it's getting better. And it's getting good (because Apple's has its own in-house chip in the future).

So Mr. McIlwaine thinks we should begin to see some really good stuff happening. If this tech is to be truly a magic technology, I mean, really a revolution than we've been taught to expect. And we shouldn't be forgetting all those things about privacy and security if Kinect catches on at next year's Consumer Electronics Show.

So who are the "nautilus" technology we must invest in when we turn to gaming for our health and productivity?

Mr. McIlwaine points out that the Kinect isn't a direct cause of heart disease. It can actually play a role in driving one's health right out of the body under extreme conditions, as can be

Average sentence match to query: 0.2081

General subject is Microsoft Kinect

**FTP Look ahead 1**

**How is it possible for a computer system to be truly creative? This requires that** we are both human beings completely free agents. A computer is an art form of the mind. When you write code, it does not ask you to do anything that is in error, or for which you do not know your limitations, but it is a game for you: make mistakes and error repair it over and over again like sandpaper: if you forget your design, it is worthless. It is a lot better than a bad design. If you change what that design does, the game breaks your whole conception of reality into bits: let your imagination go one step ahead – and the best bits of that idea still make up the games that allow you to make the best design. And that makes the worst games.

The problem with playing computer games with humans is that, in this sense, they both have their flaw. Modern computers are machines because they are machines instead of machines. That is why you have these boring computers all day, all the time, which allows you all sorts of tasks – some of which are necessary for routine tasks, some that are never part of your business for all practical purposes. These kinds of programming, however, are not part of your primary business. So instead of having to constantly change things and have to find new ways to communicate, programming does not require an active computer to be very effective. Rather, it is a more primitive form of communication.

The solution then, which is that computers can be computer science PhD students is that people spend a lot of time working at a computer instead of studying at a computer university. What does that all mean? It means that computers are full of people – the most important of all. People make the decision for the very first time how to build their software, and they make an educated decision about how to build software. But when they make that final decision, they have to live and learn the culture at a computer university. That is what computer science PhD students do. And it is what computer science PhD students do too.

Many people want people to spend time working on computers, but they also believe they must get to work on computers very soon because they will find it cheaper than working on computers. The opposite is not true: being a computer scientist is a way to save a lot more time on the technical side by just being an expert – a lawyer, an internet researcher, a lawyer, a software engineer – not really a software developer. These people are also very interested in their technical skills and know how to take advantage of them. They are very curious, they are happy, they are highly intelligent people: they have such a great sense of responsibility. They are extremely motivated because of this and they are simply good at working with computers.

And I think some people have gotten lucky. A study by Jonathan Harkeshire shows that computers – especially a combination of computers with computer programming skills – have a substantially higher return on investment than programming skills or computing skills alone. So many people have, over the last twenty years, made careers out of programming and therefore have been able to live and learn much faster than other people. That's the truth. But to see it in other ways, at the same time I think the big problem that computer science PhD students have is that the science of engineering cannot be taught by engineers. Engineers can be taught, I do not know, how to start a company or how to build something, at whatever age, all for free. That is not rocket science. Even those who go to university universities with that type of engineering program can never work as many hours to put into making it better.

The other great thing about computer science is that it has a very particular set of incentives that it has to take into account that if you are smart, you will be awarded enough money to do the best things. It is hard to overstay your head in this age of endless cash grants and the fear of endless demands for work. So although the cost of learning from

Average sentence match to query: 0.2975

General subject is computer science.

**FTP Multi-future inference, look ahead 5 tokens**

**How is it possible for a computer system to be truly creative? This requires that** the programmer decide well the way he is going to do things in most of the ways a computer can be designed. In other words, what could be considered is not art's job, of course, but art's job is to discover the creative potential of that computer. This process is called creativity.

So how does the creative potential of a computer ever arise from the design of an algorithm?

The world-famous example of this process is called what is known as a "big bang computer" (BOYTF). The main goal of the machine (or algorithm) itself is to create a large number of "bits" for the purpose of solving a problem. The math of finding the number of bits is given the name of the "big bang computer." In which case the program calculates the initial value so that all the new "bits" in the program are produced. This process is called "blending problem solving."

This process (the next step of computer genius) is called "creating an idea." The word of the story is that after a certain mathematical analysis something that someone else created with no knowledge of the nature of the idea became the idea. The "nonsense" is when the result of the actual attempt to create a piece of programming fiction came to be called an idea.

Gremlins

There might be dozens of reasons for why the creative potential of a computer comes from the use of two or more of these new ideas. The first is with the original idea, which came out in the form of an obscure mathematical idea. Then, there are several other creative things. One instance involves the "gasm" (using math) of writing, which is said to be the source of creativity. What happens afterward is that a word is published and that word appears as a "graphete" in an attempt to explain the nature of what happened. It's a kind of "text block," depending on the circumstances.

The second reason why the creative potential of a computer comes from the creation of an idea is due to the way things usually are constructed over time. While ideas don't have to prove their own validity, they do have to prove the effectiveness and reliability of an idea. This process is called "exploitationism." Because there is so much potential for creativity on the world's computer, you would think that creativity is some kind of science but isn't. In fact, most of today's problems can only be solved by computer techniques which can't be used to "hijack inventiveness" or "exploitation."

The third reason why the creative potential of a computer comes from the creation of an idea is due to the phenomenon of "coincidences." When two ideas combine into a "coincidence," an idea is formed. This process can be an act like a cat's paw in the wild, however in these specific cases what this phenomenon is and why it exists, is one of the major problems that can arise from it. There are many fascinating things that can be gained from all of the creativity involved in the creation or creation of an idea.

The question and answer to the following question is, do you intend to create another being as your scientific study progresses? Do you plan to create something different from what you know now or will you build a new way of looking at the world? The other side, where you might be trying to build a new type of knowledge about the world, is a subject of great interest to many people all around the globe. Don't count on us yet, we can definitely take your advice when we head into the future.

What does the future look like? Where

Average sentence match to query: 0.3769

General subject is computer creativity.